\documentclass[runningheads]{llncs}

\usepackage{eccvabbrv}

\usepackage{graphicx}
\usepackage{booktabs}
\usepackage{subcaption}

\usepackage[accsupp]{axessibility}  

\usepackage{hyperref}

\usepackage{orcidlink}

\begin{document}

\title{MK-ResRecon: Multi-Kernel Residual Framework for Texture-Aware 3D MRI Refinement from Sparse 2D Slices} 

\titlerunning{Abbreviated paper title}


\author{Prajyot Pyati\inst{1} \and
Sapna Sachan\inst{1} \orcidlink{0009-0001-5486-4627} \and
Amulya Kumar Mahto\inst{1} \orcidlink{0000-0001-8389-5257}\and 
Pranjal Phukan\inst{2} \orcidlink{0000-0001-9210-0217}}

\authorrunning{F.~Author et al.}

\institute{Indian Institute of Technology, Guwahati \email{akmahto@iitg.ac.in}\and
All India Institute of Medical Sciences, Guwahati \email{pranjalphukan@aiimsguwahati.ac.in}\\}

\maketitle
\begin{abstract}
Magnetic Resonance Imaging (MRI) acquisition remains a time-intensive and patient-straining process, as prolonged scan durations increase the likelihood of motion artifacts, which degrade image quality and frequently require repeated scans. To address these challenges, we propose a novel framework with two models MK-ResRecon and IdentityRefineNet3D to reconstruct high-fidelity 3D MRI volumes from sparsely sampled 2D slices-requiring only 12.5\% of the axial slices for full resolution 3D reconstruction. MK-ResRecon predicts missing intermediate 2D slices using a multi-kernel texture-aware loss, preserving fine anatomical details. IdentityRefineNet3D refines the predicted slices and the original sparse slices as a single 3D volume to obtain a smooth anatomical structure. We train the models on a large T1-sequence POST-contrast brain MRI dataset and evaluate on a large heterogeneous brain MRI cohort. The work provides accurate, hallucination-free, generalizable and clinically validated framework for 3D MRI reconstruction from highly sparse inputs and enables a clinically viable path towards faster and more patient-friendly MRI imaging.
\keywords{Magnetic Resonance Imaging \and 3D Reconstruction \and Sparsely Sampled Slices}
\end{abstract}

\section{Introduction}
\label{sec:intro}
MRI is extensively used to visualize internal anatomical structures due to its non-invasive nature. Different MRI sequences, such as T1-sequence, T2-sequence, and FLAIR are used for complementary purposes in the diagnosis of brain anatomy and pathology~\cite{haacke1999magnetic}. Brain MRI is pivotal in diagnosing tumors, lesions and other neuro-degenerative diseases. But avoiding motion artifacts becomes difficult, particularly for the pediatric and geriatric population due to long scan durations and often results in repeated scans. Furthermore, storing and processing large 3D MRI datasets imposes significant computational and memory overhead.

Traditional 3D MRI reconstruction relies on interpolation-based techniques such as linear, cubic, or spline interpolation~\cite{chung2018slice}, which are simple and fast, but are highly sensitive to slice spacing, orientation, and anatomical variability. These limitations often yield blurred or structurally inconsistent reconstructions. Recent deep learning approaches include 3D convolution networks, autoencoders, and generative adversarial networks (GANs) to learn complex spatial correlations and recover high-fidelity volumes~\cite{chartsias2017adversarial,mccann2019generative,chen2020brain}. Despite their success in various fields, they fail to maintain inter-slice consistency, prevent hallucinated structures, and minimize computational cost. The limitations pose a significant challenge in medical imaging, making it unusable in clinical deployment.
\subsection{Proposed approach}
To address these challenges, we propose a novel framework with two models \textit{MK-ResRecon} and \textit{IdentityRefineNet3D} to reconstruct high-fidelity 3D MRI volumes from sparsely sampled 2D slices. Our key objectives are to (1) predict missing slices while preserving local texture and global structure, (2) ensure volumetric consistency across slices, and (3) achieve substantial scan-time reduction without compromising diagnostic quality. To achieve the same, our work employs two stage framework consisting of,
\begin{enumerate}
    \item \textbf{MK-ResRecon: Texture-Preserving Slice Synthesis.}  
The model predicts the middle slice from two sparsely spaced inputs, (e.g., generating a slice $i$ from slices $i{-}4$ and $i{+}4$). Deploying the model recursively, we double the number of slices each time. A multi-kernel texture-aware loss along with the \(L_1\) loss enforces both structural alignment and fine-detail preservation. Attention module helps the model to focus on the anatomical regions of the image while ignoring the plain background. A hierarchical interpolation strategy enables efficient multi-level synthesis under variable sparsity.

\item \textbf{IdentityRefineNet3D: Volumetric Refinement and Consistency.}  
The sparse 2D slices along with the intermediate slices predicted by \textit{MK-ResRecon} are integrated as a 3D volume and refined using \textit{IdentityRefineNet3D}, which is a Light weight 3D CNN module that improves volumetric continuity and suppresses hallucinations. The identity-preserving residual path preserves the structure of the anatomy, while a multi-kernel 3D~$L_1$ loss (using Sobel, Laplacian and diagonal kernels) promotes texture- and structure-aware optimization. The depth-sensitive weighting of the slices further emphasizes the sparse original input slices.
\end{enumerate}
\subsection{Novelty}
The most recent papers focus on tasks such as 2D super-resolution ~\cite{yang2025_transformer_sr}, reconstruction on $\kappa$ -space data ~\cite{bangun2025mri,chung,chung2020} ,and 3D super-resolution ~\cite{zhang2021_2d_to_3d_sr} and have shown very good results. Since the MRI acquisition time is directly proportional to the number of slices in the depth direction, our work specifically focuses on \textbf{reconstruction of full 3D volume using few number of slices in depth direction directly in the image space}. The novelty of the methodology lies in the multiple components, each one specifically designed for a specific task such as:
\begin{itemize}
    \item \textbf{Framework:} The work is among the firsts to employ recursive 2D slice interpolation followed by 3D refinement step. The framework was specifically designed to mitigate hallucination.
    \item \textbf{MK-ResRecon:} The MK-ResRecon is the model specifically designed to interpolate the missing middle slice.
    \item \textbf{Multi-Kernel Loss:} A loss function which uses multiple kernels to give us more control over the individual features of the generated image.
    \item \textbf{IdentityRefineNet3D:} Extra-light refinement model designed to enhance the continuity and smoothness of the 3D volumes without altering the structural consistency.
    \item \textbf{Clinical Validation:} Expert evaluation of the hallucination free claim of the work adds to the applicability of the work.
\end{itemize} 

\section{Related Work}
\noindent
\textbf{Interpolation and model-based reconstruction: }Early 3D MRI reconstruction methods estimate missing slices using linear, cubic, or spline interpolation~\cite{borse2013literature,chung2018slice}. These approaches are computationally inexpensive but highly sensitive to slice spacing and anatomical variability, leading to blurred and inconsistent volumes. Population-based imputation methods, such as Dalca~\textit{et al.}~\cite{dalca2017_population_imputation}, generate plausible anatomical structures from sparse clinical scans but rely on large population datasets and exhibit limited cross-institution generalization. Frame-interpolation-based slice completion methods~\cite{wu2022_interpolation} improve segmentation performance on anisotropic volumes but fail to capture fine details and subtle pathologies.

\medskip
\noindent
\textbf{Deep learning for slice synthesis and super-resolution: }Deep CNNs and GAN-based methods have advanced MRI reconstruction by learning non-linear mappings from sparse to dense slices. Zhang~\textit{et al.}~\cite{zhang2021_2d_to_3d_sr} proposed a two-stage super-resolution pipeline (RFB-ESRGAN + nESRGAN) to upsample in-plane slices and synthesize missing ones, but encountered issues with inter-slice brightness variation and checkerboard artifacts. Remedios~\textit{et al.}~\cite{remedios2023_selfsr} introduced a self-supervised SR approach that mitigates data scarcity yet depends heavily on contrast-specific anatomical features. Bilgic~\textit{et al.}~\cite{bilgic2020_2d_to_3d} reconstructed isotropic 3D FLAIR from thick 2D slices, achieving perceptual improvements but limited adaptability across varying slice thicknesses. Despite improved sharpness, most 2D models reconstruct slices independently, resulting in poor volumetric coherence. 

\medskip
\noindent
\textbf{Self-supervised and transformer-based reconstruction: }Transformers with cross-slice attention~\cite{yang2025_transformer_sr} capture inter-slice dependencies efficiently but remain limited in modeling long-range 3D continuity and are sensitive to artifact propagation. Self-supervised SR frameworks~\cite{remedios2023_selfsr} enhance data efficiency but depend on the anatomical priors of the training distribution, hindering generalization to unseen modalities or pathologies.

\medskip
\noindent
\textbf{Physics-aware and model-based deep learning: }Model-based deep learning integrates physics priors with neural optimization for MRI reconstruction. Methods like E2E-VarNet~\cite{sriram2020_e2evarnet} and MoDL~\cite{sriram2020_e2evarnet} combine data-consistency with CNN regularization, achieving state-of-the-art performance on fastMRI~\cite{zbontar}, but rely on multi-coil data, limiting their applicability to image-domain sparse-slice tasks. Diffusion-based approaches such as ScoreMRI~\cite{chung} and TPDM~\cite{lee} leverage generative diffusion priors for improved structural fidelity, while X-Diffusion~\cite{bourigault2024x} and R3DM~\cite{bangun2025mri} extend these methods to volumetric reconstruction with enhanced anatomical consistency. However, these models remain computationally demanding and may exhibit minor inter-slice inconsistencies, posing challenges for clinical scalability.

\medskip
\noindent
\textbf{Multi-modal and attention-enhanced networks: }ARD-U-Net~\cite{yang2024_ardunet} employs residual dense blocks and dual-branch squeeze-and-excitation attention to fuse multi-modal MRI inputs for enhanced reconstruction fidelity. While successful for 2D cross-modal recovery, it requires fully sampled auxiliary modalities, which is an additional constraint on the applicability of the model and it also entails high computational cost. Such 2D designs still lack volumetric enforcement needed for consistent 3D synthesis.

\medskip
\noindent
\textbf{Summary and Research Gap: }
The existing methods face persistent challenges: (1) \textit{inter-slice inconsistency} due to 2D processing; (2) \textit{hallucination artifacts} from adversarial or diffusion synthesis; (3) \textit{high computational demands} of transformer and generative approaches; and (4) \textit{limited generalization} across anatomy and contrast domains. To address these, we propose \textbf{MK-ResRecon}, a new paradigm combining texture-preserving slice synthesis and volumetric refinement under a unified framework. Through multi-kernel residual learning, texture-aware optimization, and depth-weighted 3D refinement, MK-ResRecon delivers high-fidelity volumetric reconstruction while substantially reducing acquisition time—bridging the gap between fast MRI acquisition and clinical-grade image quality.

\section{Dataset Description}
We consider a curated subset of the \textit{Yale-Brain-Mets-Longitudinal} dataset from The Cancer Imaging Archive (TCIA)~\cite{chadha2025yale}, containing longitudinal brain MRI studies of patients with clinically confirmed brain metastases. The dataset being longitudinal contains both healthy and pathological brain MRI volumes. Our subset focuses on post-contrast T1-weighted (T1c) volumes, comprising of 710 3D scans ($126{,}339$ axial slices) for training and 80 scans ($14{,}231$ slices) for testing. In addition to this, 259 3D scans from FLAIR, T1-pre, and T2-weighted sequences are included to evaluate multi-modal generalization. To ensure high spatial fidelity, only scans having slice thickness less than 1.5 mm are retained. As discussed above, the dataset includes both healthy and pathological cases, and its fine slice spacing facilitates accurate reconstruction using sparsely sampled slices—aligning with our objective of reducing MRI acquisition time.
For quantitative benchmarking, we also utilize the \textit{BraTS 2020} dataset~\cite{bakas,bakas2018,menze}, containing 5,880 skull-stripped MRI scans from 1,470 glioma patients. All volumes are preprocessed to 1 mm isotropic resolution and standardized to $240\times240\times155$. We randomly select 100 FLAIR T2 volumes from BraTS for testing to assess the performance of reconstruction  and cross-dataset generalization.

\section{Methodology}
\label{sec:method}
\begin{figure*}[t]
    \centering
    \includegraphics[width=0.98\textwidth]{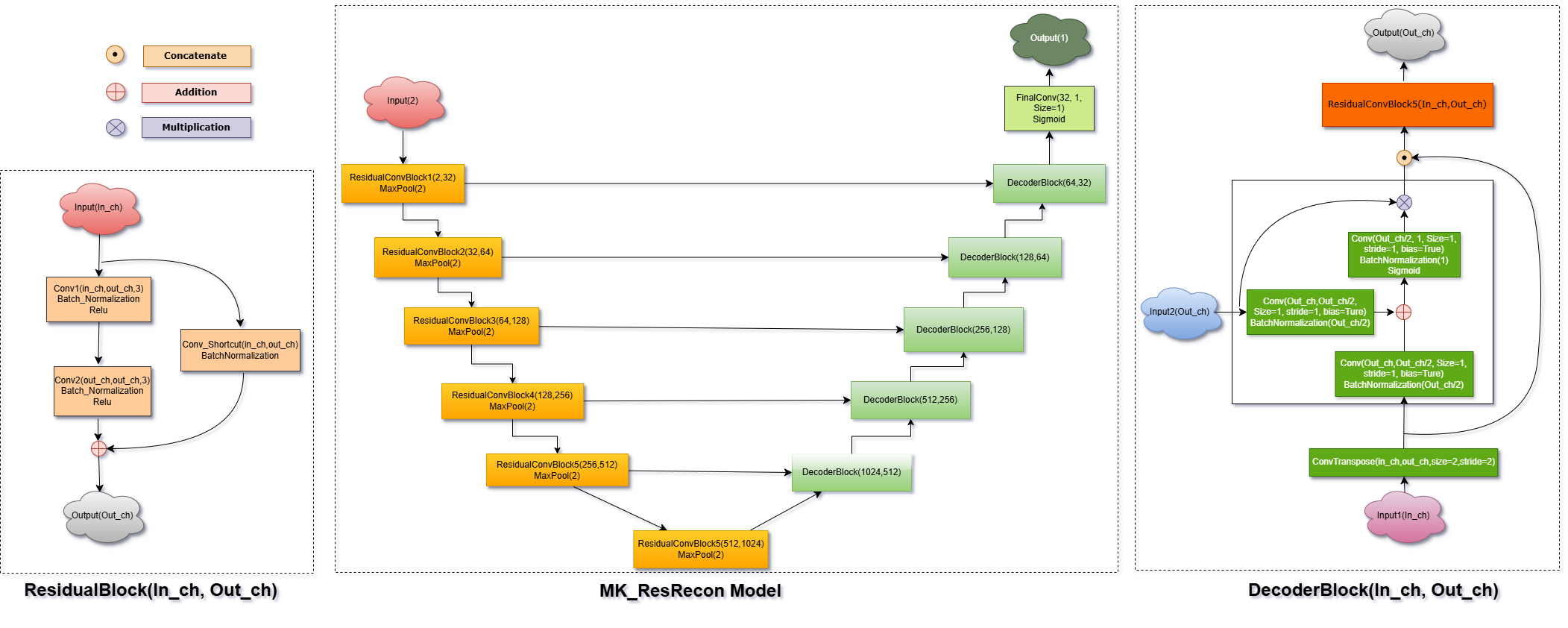}
    \caption{
        \textbf{Stage~1 architecture for 2D slice synthesis.}
        Two non-adjacent slices are input to generate intermediate slices using attention-residual connections for edge and texture fidelity.
    }
    \label{fig:stage1_arch}
\end{figure*}
The framework operates in two complementary stages.
\subsection*{Stage 1: 2D Missing-Slice Synthesis}

In Stage~1, an \textbf{Attention-Residual U-Net} synthesizes intermediate 2D slices between widely spaced acquisitions.
The model is recursively applied to reconstruct slices for varying gaps (2, 4, and 8), thereby assessing robustness under different acquisition sparsity levels. An overview of the stage~1 architecture is provided in Fig.~\ref{fig:stage1_arch}.

The features of the proposed model include:
\begin{itemize}
    \item \textbf{Residual learning:} It is helpful for the task of predicting the missing slices, where the input is very similar to the output.
    \item \textbf{Attention gates:} It helps to focus on important structures ignoring background noise.
    \item \textbf{Multi-Kernel loss:} It enables the model to translate the features (for example: edges, textures and gradients) with weighted importance.
\end{itemize}

The total loss for Stage~1 combines pixel-level fidelity and local texture-preserving consistency:
\begin{equation}
\mathcal{L}_{\text{total}} =
\lambda_1 \, \mathcal{L}_{\text{$L_1$}} +
\lambda_2 \, \mathcal{L}_{\text{filtered}},
\end{equation}
\noindent where $L_{\text{filtered}}$ is a multi-kernel 2D $L_1$ loss and $(\lambda_1, \lambda_2)$ loss weights balance global and local reconstruction quality.


\begin{figure*}[t]
\centering

\begin{minipage}[h]{0.45\textwidth}
\centering
\includegraphics[height=4.8cm]{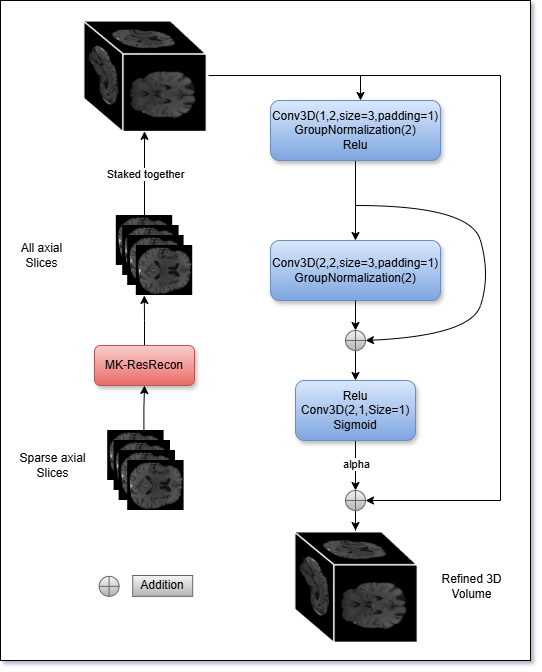}
\captionof{figure}{The overview of stage~1 and Stage~2 architecture (IdentityRefineNet3D) for volumetric refinement. The network refines the stacked 3D volume using residual identity blocks.}
\label{fig:stage2_arch}
\end{minipage}
\hfill
\begin{minipage}[h]{0.5\textwidth}
\centering
\captionof{table}{Every model takes two slices with slice gap of 8 as inputs (conditions for conditional diffusion-based models) and gives the middle slice as output.}
\label{tab:recomp_1}
\begin{tabular}{l c c}
\hline
\textbf{Model} & \textbf{PSNR} & \textbf{SSIM}  \\
\hline
X-Diffusion & 28.54 & 0.5302 \\
UNet & 30.94 & 0.9013 \\
MedGAN & 31.54 & 0.8969 \\
Pix2Pix & 32.21 & 0.9017 \\
MedDiffusion & 33.14 & 0.8657 \\
ResViT & 34.86 & 0.9381 \\
SWinUNET & 34.93 & 0.9386 \\
MedViT & 35.49 & 0.9457 \\
\hline
\textbf{MK-ResRecon (ours)} & \textbf{37.43} & \textbf{0.9636} \\
\hline
\end{tabular}
\end{minipage}
\end{figure*}

\subsection*{Stage 2: 3D Volumetric Refinement}

The synthesized and real slices are stacked to form a coarse 3D volume $\mathcal{I}$, which is the input of \textbf{IdentityRefineNet3D} — a lightweight model designed that it returns the refinements for volumetric smoothness, suppressing residual artifacts, and restoring anatomical continuity.\newline An overview of the pipeline is illustrated in Fig.~\ref{fig:stage2_arch}.\newline\textbf{IdentityRefineNet3D} returns the refinement $\mathcal{R}$ which is scaled down and added to the original 3D volume to obtain the final output

Predicted image $\mathcal{P}$ is calculated as follows.

\begin{equation}
\mathcal{P} = \mathcal{I} + \alpha \, \mathcal{R},
\end{equation}

\noindent where $\alpha$ is a small weighting factor that prevents drastic modifications to the input.

Key advantages of the approach:
\begin{itemize}
    \item \textbf{A lightweight model} minimizes overfitting and reduces the risk of hallucinated structures
    \item \textbf{Identity preservation:} Since $\alpha$ is a small number, the model is restricted to do major anatomical changes.
\end{itemize}

The Stage~2 loss integrates voxel-wise accuracy and texture-aware volumetric consistency:
\begin{equation}
\mathcal{L}_{\text{total}} =
\alpha_1 \, \mathcal{L}_{\text{$L_1$}} +
\alpha_2 \, \mathcal{L}_{\text{filtered}},
\end{equation}
\noindent
where $L_{\text{filtered}}$ is a multi-kernel 3D $L_1$ loss and $(\alpha_1, \alpha_2)$ is loss weights to balance structure preservation and fine-texture refinement.



\subsection*{Multi-Kernel Loss Functions}
\label{subsec:filteredloss}
The existing loss functions($L_1$ or MSE) work flawlessly in various applications. The losses penalize the dominant features(boarders, thick lines ect) of the two images very well, but fail to consider the finer details especially when the images are too complex.  
To address this and enhance recovery of high-frequency anatomical details such as cortical boundaries and tissue textures, we introduce the \textbf{Multi-kernel $L_1$ Loss}, extended for both 2D and 3D domains.

\subsubsection*{(a) Standard $L_1$ Loss}

The standard $L_1$ loss between predicted image $\mathbf{P}$ and target image $\mathbf{T}$ is defined as:
\begin{equation}
\mathcal{L}_{\text{$L_1$}} = 
\frac{1}{|\Omega|}
\sum_{v \in \Omega} 
\big| \mathbf{P_v} - \mathbf{T_v} \big|,
\end{equation}
\noindent
where $\Omega$ is the set of all pixels(voxels) co-ordinates and $|\Omega|$ is total number of pixels(voxels) 

\subsubsection*{(b) Multi-kernel $L_1$ Loss}

Let $\{F_k\}_{k=1}^{K}$ denote a set of spatial kernels (Sobel-X/Y, Laplacian, diagonal and checkerboard kernels) and center $c$.  
The multi-kernel $L_1$ loss that emphasizes differences in local gradients and textures is defined as:
\begin{equation}
\mathcal{L}_{\text{Filtered}} =
\sum_{k=1}^{K} w_k \frac{1}{|\Omega|}
\sum_{v\in\Omega}
\Big| (\mathbf{P} * F_k)_v - (\mathbf{T} * F_k)_v \Big|,
\end{equation}
\noindent where $\Omega$ denotes all pixel(voxel) co-ordinates, and $|\Omega|$ is total number of pixels(voxel) $\{w_k\}$ are normalized weights i.e. $\sum_{k=1}^{K} w_k = 1$ and $\mathbf{P} * F_k$ represents the operation defined as: 
\begin{equation}
(P*F_k)_{v \in \Omega}
\;=\; \sum_{u \in \omega}
F_k[u] \; P[v+(u-c)].
\end{equation}
\noindent
where $\omega$ is the set of coordinates of the kernel $F_k$.

This loss function gives us more control over the features of the images that we want to focus on. The weight $w_k$ of the Kernel $F_k$ is the amount of penalty assigned to that particular feature. This loss penalizes discrepancies in multi-directional gradient responses, enabling sharper 3D reconstructions with anatomical consistency.

\subsection*{Training Configuration}
\paragraph{Stage 1.}  
Trained using the Adam optimizer with initial learning rate $1\times10^{-4}$, batch size = 8 with ReduceLRonPlatue scheduler (mode='max', patience=5, factor=0.5) based on validation PSNR value.
Data processing includes image resizing to 256$\times$256 and intensity normalization to [0,1] to improve generalization.
 Loss weights are set as $(\lambda_1, \lambda_2) = (0.1, 1.0)$. The multi-kernel loss function uses the multiple kernels(sobel, diagonal, laplacian and checkerboard). The exact kernels and weights we got experimentally are mentioned in the supplementary information. 
\paragraph{Stage 2.}  
The 3D refinement network is trained independently for 50 epochs using Adam ($\text{lr}=10^{-4}$) and batch size = 4.
Unlike the first stage the Loss weights are not fixed. $\alpha_1$ is fixed as $\alpha_1=0.25$ but $\alpha_2$ is initialized as $\alpha_2=1$ then with each epoch, the $\alpha_2$ decreases by 0.016. The multi-kernel loss function uses the multiple 3D kernels(sobel, diagonal, laplacian and checkerboard). The exact kernels and weights we got experimentally are also mentioned in the supplementary information. 
\section{Evaluation Metrics}
\label{sec:metrics}

To quantitatively assess the quality of reconstructed axial slices and volumetric MRI data, we employ two complementary image fidelity metrics: \textbf{PSNR} and \textbf{SSIM}.

\subsection*{Peak Signal-to-Noise Ratio (PSNR)}
PSNR measures the global similarity between a reconstructed image $\mathbf{P}$ and the reference ground-truth image $\mathbf{T}$ in terms of intensity fidelity. It is defined as:
\begin{equation}
\text{PSNR} = 10 \cdot \log_{10} \left( \frac{L^2}{\text{MSE}} \right),
\end{equation}
where 
\begin{equation}
\text{MSE} = \frac{1}{|\Omega|} \sum_{i \in \Omega} \big( \mathbf{P}_{i} - \mathbf{T}_{i} \big)^2,
\end{equation}
where $\Omega$ is set of all pixel (voxel) co-ordinates
and $L$ denotes the maximum possible pixel intensity value (for normalized MRI images, $L = 1$).
\subsection*{Structural Similarity Index Measure (SSIM)}
SSIM captures perceptual and anatomical consistency by comparing luminance, contrast, and structure between $\mathbf{P}$ and $\mathbf{T}$.  
It is defined as:
\begin{equation}
\text{SSIM}(\mathbf{P}, \mathbf{T}) =
\frac{(2\mu_{\mathbf{P}}\mu_{\mathbf{T}} + C_1)(2\sigma_{\mathbf{P}\mathbf{T}} + C_2)}
{(\mu_{\mathbf{P}}^2 + \mu_{\mathbf{T}}^2 + C_1)(\sigma_{\mathbf{P}}^2 + \sigma_{\mathbf{T}}^2 + C_2)},
\end{equation}
where $\mu_{\mathbf{P}}$, $\mu_{\mathbf{T}}$ are mean intensities, $\sigma_{\mathbf{P}}^2$, $\sigma_{\mathbf{T}}^2$ are variances, and $\sigma_{\mathbf{P}\mathbf{T}}$ is the covariance between the two images.  
Constants $C_1$ and $C_2$ stabilize the division for small denominators. Here, we choose $C_1=10^{-4}$ and $C_1=9 \times 10^{-4}$.
\section{Experimental Results and Analysis}

\label{sec:results}
\begin{figure*}[h]
    \centering
    \includegraphics[width=0.8\textwidth]{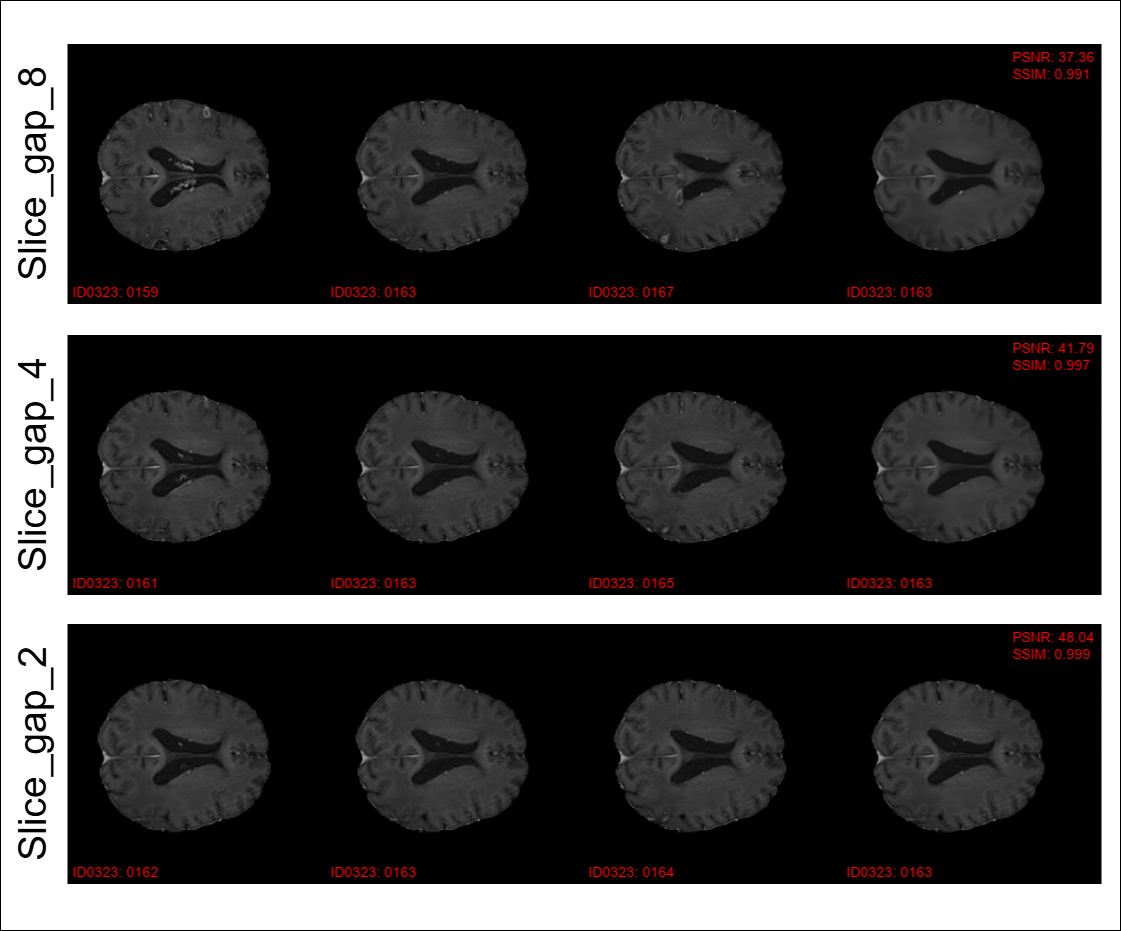}
    \caption{
\textbf{2D axial reconstruction performance across varying slice gaps.}
Illustrated are representative 2D axial examples (e.g., patient ID0323), where slice 0163 is reconstructed from its neighboring slices 0159 and 0168 (\textbf{slice gap = 8}). 
The figure compares reconstruction quality for slice gaps of 2, 4, and 8, demonstrating the model’s robustness to different levels of acquisition sparsity. 
Each row corresponds to a specific slice gap—(top) 8, (middle) 4, and (bottom) 2—showing the generated slice alongside its adjacent input slices. 
Quantitative measures (PSNR and SSIM) are annotated in the top-right corner of each panel. 
As the slice gap increases, reconstruction fidelity decreases gradually, yet the model maintains high structural consistency and anatomical coherence under sparse input conditions.
}
\label{fig:slice_gap_comparison}
\vspace*{-1mm} 
\end{figure*}
\subsection{Quantitative Evaluation and Baseline Comparison for 2D Slice Reconstruction}
To establish quantitative baselines for the proposed framework, we first evaluate the 2D slice generation task—reconstructing missing slices from sparsely sampled 2D MRI sequences.
Recent generative models—such as GAN-based and diffusion-based priors~\cite{chung,zhang2021_2d_to_3d_sr}—have demonstrated strong synthesis capabilities in medical image reconstruction. However, these approaches typically rely on dense or fully-sampled inputs, adversarial objectives, and modality-specific training, which differ fundamentally from our sparse-slice reconstruction setup. Moreover, they often introduce hallucinated or non-existent anatomical structures, compromising clinical reliability. Similarly transformer-based architectures~\cite{yang2025_transformer_sr} effectively capture inter-slice dependencies but remain limited in enforcing consistent 3D volumetric continuity under high sparsity.

Due to the limited availability of prior work explicitly addressing this sparse-to-dense reconstruction setting, we compare against conventional deep learning baselines: CNN , Transformer, Diffusion, and GAN based architectures. Table~\ref{tab:recomp_1} reports the corresponding PSNR and SSIM values for 8 slice gap. The experimental results show that transformers perform best among the baseline exceeding the GANs and diffusions.

Our \textbf{MK-ResRecon} framework achieves the superior PSNR(37.43 dB) and SSIM(0.9636) showing balanced trade-off between quantitative accuracy and structural fidelity for an 8-slice gap while maintaining hallucination-free reconstructions consistent with true anatomical structures.

\subsection{Qualitative Assessment of 2D Slice Reconstruction}
Figure~\ref{fig:slice_gap_comparison} presents qualitative 2D axial reconstruction results under varying slice gaps (\(2\), \(4\), and \(8\)). Each example illustrates how the proposed method preserves edge sharpness, tissue textures, and structural integrity even when only a fraction of the original slices are available. As the slice gap increases, a gradual decrease in quantitative fidelity is observed. However, the reconstructed slices maintain high perceptual similarity to the ground truth, validating the model’s ability to generalize across different acquisition sparsity levels. The visual consistency across gaps highlights the effectiveness of our multi-kernel loss in preserving both global structure and local texture continuity. Detailed visual results are given in the supplementary section that represents the size, location and intensity of the lesions and other anatomical features in ground truth images and generated images.

\subsection{3D Volumetric Refinement and Generalization Analysis}
Tables~\ref{tab:metrics} and~\ref{tab:metrics_unknown} report quantitative results for 3D MRI reconstruction under different slice-gap settings. Table~\ref{tab:metrics} presents results on the \textbf{T1 Post-Contrast} test set, which is the modality used during training, while Table~\ref{tab:metrics_unknown} evaluates performance on \textbf{unseen modalities} (FLAIR, T2, and T1 Pre-Contrast), demonstrating the model’s generalization capability.
\begin{table}[h!]
\centering
\caption{Quantitative evaluation of reconstructed \textbf{ T1 Post-Contrast images for 3D MRI} with and without refinement.}
\label{tab:metrics}
\setlength{\tabcolsep}{4pt}
\renewcommand{\arraystretch}{1.0}
\small
\begin{tabular}{c c l}
\toprule
\textbf{Slice-Gap} & \textbf{With Refinement}  & \textbf{Without Refinement}\\
\midrule
{2} & PSNR: 49.8851 dB & PSNR: 51.1813 dB \\
                   & SSIM: 0.9982 & SSIM: 0.9954 \\[3.5pt]
\hline                   
{4} & PSNR: 42.1984 dB & PSNR: 43.3284 dB \\
                   & SSIM: 0.9928 & SSIM: 0.9789 \\[3.5pt]
\hline                   
{8} & PSNR: 38.0991 dB & PSNR: 39.3355 dB \\
                   & SSIM: 0.9828 & SSIM: 0.9563 \\
\bottomrule
\end{tabular}
\end{table}
\begin{table}[h!]
\centering
\caption{Quantitative evaluation of reconstructed \textbf{ testing on unseen (multi-modal—FLAIR, T2, T1 Pre-Contrast) 3D MRI} with and without refinement.}
\label{tab:metrics_unknown}
\setlength{\tabcolsep}{10pt}
\renewcommand{\arraystretch}{1.3}
\small
\begin{tabular}{c c l}
\toprule
\textbf{Slice-Gap} & \textbf{With Refinement} & \textbf{Without Refinement} \\
\midrule
{2} & PSNR: 49.7706 dB & PSNR: 51.4049 dB \\
                   & SSIM: 0.9986 & SSIM: 0.9966 \\[3.5pt]
\hline                   
{4} & PSNR: 41.3939 dB & PSNR: 42.7792 dB \\
                   & SSIM: 0.9934 & SSIM: 0.9771 \\[3.5pt]
\hline                   {8} & PSNR: 36.7020 dB & PSNR: 37.7673 dB \\
                   & SSIM: 0.9822 & SSIM: 0.9508 \\
\bottomrule
\end{tabular}
\end{table}Here, \textbf{with refinement} refers to the final 3D reconstruction after applying the refinement stage, which enhances volumetric smoothness and inter-slice continuity. In contrast, \textbf{without refinement} denotes the coarse 3D volume obtained by directly stacking the synthesized and original slices, without any post-processing. Note that PSNR only concerns about the voxel level accuracy and penalizes any deviation even perceptually irrelevant ones. In contrast, SSIM measures the similarity in terms of structure, luminance, and contrast. Higher SSIM indicates visual realism, sharp edges, and texture continuity, and tolerates small voxel deviations if the structure is preserved. Since the 3D Multi-Kernel loss is specifically designed to preserve structural details, the refinement stage focuses more on structural continuity rather than voxel level accuracy. This explains the small drop in PSNR and the significant increase in the SSIM after the refinement stage.
Furthermore, Table \ref{tab:metrics_comp} shows the comparison of recent state of the art(SOTA) models~\cite{lee,bourigault2024x} that address the \textit{slice-to-volume} MRI reconstruction problem under sparse sampling conditions. Since the model MK-ResRecon takes two slices with perticular slice gap as input and predicts the middle slice and MK-ResRecon is applied to the input recursively which doubles the number of slices each time, our framework cannot produce the acceleration rate of 5(not a power of 2). Moreover, the given models show the best results on acceleration rate of 5. Hence, we compare the SOTA models with slice gap of 5 to our model with slice gap 8. Despite being trained on the Post Contrast T1 sequence data and larger slice-gap, our model performs better than the models mentioned which are trained on the T2-FLAIR sequence data which shows the effective generalizability of our model.

\begin{table}[h]
\centering
\caption{
\textbf{Quantitative comparison of reconstruction performance on 3D Brain MRI(BraTS 2020) FLAIR T2 sequence.}
}
\label{tab:metrics_comp}
\setlength{\tabcolsep}{3pt}
\renewcommand{\arraystretch}{1.2}
\small
\begin{tabular}{lcccc}
\toprule
\textbf{Model} & \textbf{Slice Gap} &  \textbf{Trained on} & \textbf{Tested on} & \textbf{PSNR(dB)} \\
\midrule
ScoreMRI~\cite{chung} & 5 & FLAIR T2 & FLAIR T2 & 29.24 \\
TPDM ~\cite{lee} & 5 & FLAIR T2 & FLAIR T2 & 31.48 \\
\textbf{MK-ResRecon (Ours)} & 8 & T1 Post-C. & FLAIR T2 & 33.24 \\
\bottomrule
\end{tabular}
\end{table}

\begin{figure}[!ht]

\end{figure}

\begin{figure*}[h!]
\centering

\begin{minipage}[h]{0.45\textwidth}
\centering
\includegraphics[width=\textwidth]{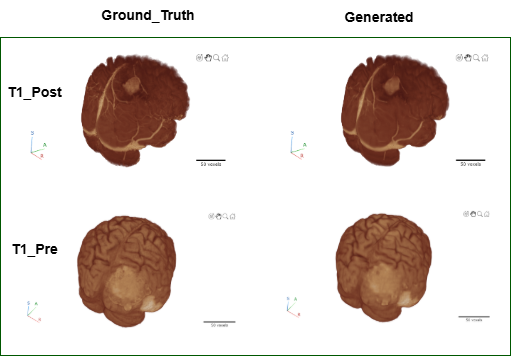}
\caption{
\textbf{Qualitative 3D reconstruction results.}
The figure compares ground-truth and generated 3D MRI volumes across two modalities. 
\textbf{Top row:} \textbf{T1 Post-Contrast} — the model accurately reconstructs tumor regions and preserves fine structural integrity. 
\textbf{Bottom row:} \textbf{T1 Pre-Contrast} — the model generalizes effectively to unseen data, producing anatomically consistent reconstructions without transferring modality-specific contrasts such as the bright enhancement lines (Venous Sinus) observed only in T1 Post-Contrast scans.
}
\label{fig:3D_results}
\end{minipage}
\hfill
\begin{minipage}[h]{0.48\textwidth}
\centering
\includegraphics[width=\textwidth]{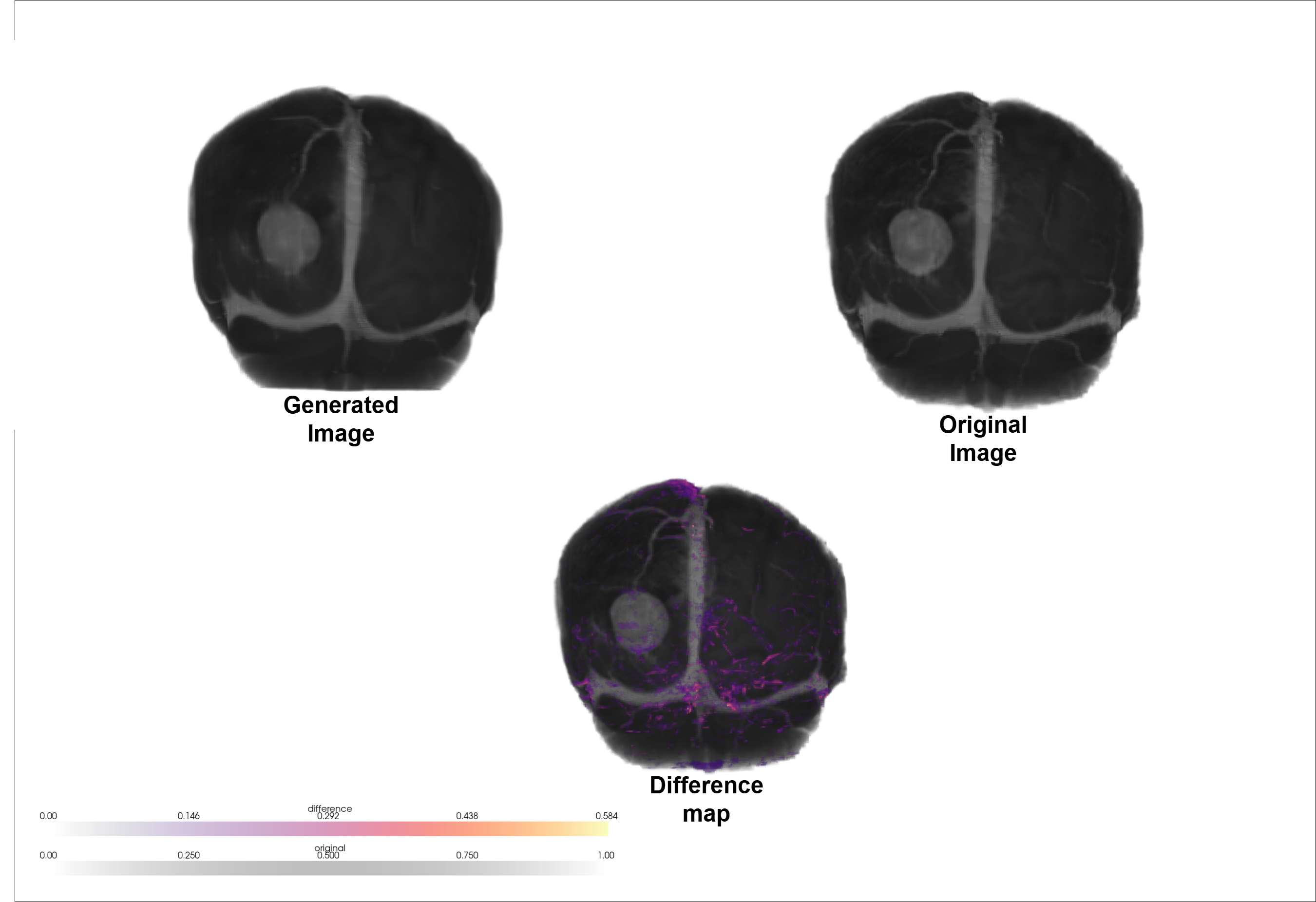}
\caption{
\textbf{Difference maps of 3D reconstructions}
The figure compares ground-truth and generated 3D MRI volumes using difference maps. The figure presents original, generated and difference map images of T1 Post Contrast modality. The difference map is the original image where the difference is highlighted.
}
\label{fig:Difference_maps}
\end{minipage}
\end{figure*}
\subsection{3D MRI Visualization and Analysis}
Figure~\ref{fig:3D_results} illustrates qualitative comparisons between the reconstructed 3D MRI volumes and their corresponding ground truth for both \textbf{T1 Post-Contrast} and \textbf{T1 Pre-Contrast}. The proposed \textbf{MK-ResRecon} framework effectively preserves tumor boundaries, tissue contrast, and fine anatomical structures, producing visually coherent and anatomically realistic 3D reconstructions.  
Unlike existing works, the reconstructions remain \textbf{hallucination-free}, avoiding the generation of artificial or non-existent structures that could compromise diagnostic reliability. As observed from Fig \ref{fig:3D_results}, the T1 Post-Contrast reconstruction accurately translates the tumor region without introducing any spurious details, while the T1 Pre-Contrast results demonstrate \textbf{strong generalization ability} of the model, faithfully reconstructing MRI volumes from unseen modalities without transferring modality-specific artifacts. Particularly, the bright enhancement regions (“white lines”) characteristic of T1 Post-Contrast scans are not mistakenly reproduced in the T1 Pre-Contrast outputs, confirming that the model learns anatomically meaningful representations rather than direct intensity mapping. More detailed and explanatory visual 3D results are mentioned and are given as supplementary material.

Figure \ref{fig:Difference_maps} presents the original, generated and absolute difference images. The difference map is the original image where the absolute difference from the generated image to the original image is colored. The difference is fairly random, and no anatomy specific differences are evident. The lesion boundary can be clearly seen in gray, showing that the lesion boundary is well preserved. This reinforces the claim of high anatomical fidelity along with the better quantitative results.

These findings demonstrate that \textbf{MK-ResRecon} not only achieves high quantitative performance but also maintains clinical consistency and interpretability across different modalities, reinforcing its suitability for reliable volumetric MRI reconstruction.

Overall, the proposed \textbf{MK-ResRecon} framework delivers high-fidelity, hallucination -free volumetric reconstruction while substantially reducing acquisition time-bridging the gap between fast MRI acquisition and clinically reliable image quality. By jointly leveraging texture-preserving synthesis and volumetric refinement, the framework establishes a unified, efficient, and anatomically consistent solution for 3D MRI reconstruction from sparse inputs.

\section{Clinical Validation and Expert Evaluation}
A highly experienced radiologist with more than 16 years of experience from a reputed medical institution/hospital has validated the 80 T1 POST C sequence images which are not used for training by comparing the generated volumes (using 8 slice gap) and the ground truth volumes on the following primary and secondary criteria. The findings of the radiologist are as mentioned below.

\subsection{Primary criteria}
\begin{enumerate}
   
\item \textbf{Pathological cases}
\begin{itemize}
    \item \textit{\textbf{Lesion morphology:}} No significant differences were observed between the lesion morphology(size, shape and location) in reconstructed and ground truth volumes.

    \item \textit{\textbf{Perilesional edema:}} The extent and morphology(size, shape and location) of perilesional edema in the reconstructed datasets matched those in the original scans, indicating accurate preservation of pathological margins.
\end{itemize}
 
\item \textbf{Healthy cases}
\begin{itemize}
    \item \textit{\textbf{Hallucinated lesions:}} None detected.
    \item \textit{\textbf{Ventricle size and orientation:}} No significant differences were noted in the dimensions of the bilateral lateral ventricles (frontal, occipital, and temporal horns) or the third ventricle between reconstructed and reference datasets.
\end{itemize}

\end{enumerate}

\subsection{Secondary criteria}

\begin{enumerate}

    \item Spurious enhancements introduced by reconstruction (e.g., inflammation or infection): \textbf{None detected.}

    \item Missed enhancements present in ground-truth images but absent in reconstructed images: \textbf{None detected.}
    
\end{enumerate}

The comparative evaluation of the model’s reconstructed volumes against the ground truth by the radiologist further substantiates the model’s reliability. This validates the hallucination-free behavior of the model and its remarkable fidelity in structural and pathological reconstruction. The close alignment between the generated and reference volumes across both healthy and pathological cases confirms that the model faithfully preserves anatomical and pathological details without introducing any spurious or missing features, thereby demonstrating its robustness and clinical applicability. Other comments of the radiologist can be found as supplementary material.

\section{Conclusion and Future work}
We presented \textbf{MK-ResRecon}, a unified framework for reconstructing dense 3D MRI volumes from sparsely sampled 2D slices. The proposed model combines a multi-kernel loss with volumetric refinement to ensure structural continuity, texture preservation, and hallucination-free synthesis. Extensive quantitative and qualitative evaluations demonstrate that MK-ResRecon outperforms conventional baselines, maintaining high anatomical fidelity even under severe slice sparsity. Building on these results, we further estimate the potential reduction in MRI acquisition time achievable through our framework, following the standard scan time formulation described in~\cite{bernstein2004handbook}.

\begin{equation}
    {Scan\_time} \propto  TR\times N\times NEX \times Phase\_Matrix    
\end{equation}
where $TR$ is repetition time, $N$ is total number of acquired slices, $Phase\_Matrix$ is the resolution of each 2D slice and $NEX$ is the number of times a single slice is sampled.

Since the MRI scan time is directly proportional to the number of acquired slices ($N$), the proposed \textbf{MK-ResRecon} framework, which reliably reconstructs missing slices with a gap of 8, effectively reduces the total scan time by a factor of approximately \textbf{8}. Considering practical acquisition constraints and patient motion factors, the achievable reduction in scan time can be conservatively estimated to be in the range of \textbf{5-6×}, without compromising diagnostic quality.

Future extensions will focus on optimizing MRI acquisition parameters, particularly slice thickness and slice spacing, to achieve an optimal trade-off between scan efficiency and reconstruction fidelity. Based on current findings, a slice thickness of approximately \textbf{0.7 mm} and a slice spacing of \textbf{5.6 mm} are recommended for sparse 2D acquisitions. Furthermore, it focuses on leveraging the domain adaptive models for different MRI machines.


%
%
\bibliographystyle{splncs04}
\bibliography{main}

\clearpage
\setcounter{page}{1}
\begin{center}
    \textbf{Supplimentary Material}
\end{center}
\section*{2D Slice Prediction}
\label{sec:Supplimentary}
The Kernels and weights for 2D Muti-kernel loss are as follows.
\begin{enumerate}
    \item \textbf{Sobel filters(Detecting straight edges)}
\vspace{0.3cm}

    Horizontal edge:
    $\begin{bmatrix}
        1 & 2 & 1 \\
        0 & 0 & 0 \\
        -1 & -2 & -1\\
    \end{bmatrix}$
\vspace{0.3cm}
    
    Verticle edge:
    $\begin{bmatrix}
        1 & 0 & -1 \\
        2 & 0 & -2 \\
        1 & 0 & -1\\
    \end{bmatrix}$

\vspace{0.7cm}
    \item \textbf{Diagonal Filters(Detecting Diagonal edges)}
\vspace{0.3cm}
    Diagonal 1:
    $\begin{bmatrix}
        0 & 1 & 2 \\
        -1 & 0 & 1 \\
        -2 & -1 & 0\\
    \end{bmatrix}$
\vspace{0.3cm}

    Diagonal 2:
    $\begin{bmatrix}
        2 & 1 & 0 \\
        1 & 0 & -1 \\
        0 & -1 & -2\\
    \end{bmatrix}$

\vspace{0.7cm}

    \item \textbf{Laplacian filters(Detecting fine lines and structures)}
\vspace{0.3cm}

    Laplacian:
    $\begin{bmatrix}
        0 & 1 & 0 \\
        1 & -4 & 1 \\
        0 & 1 & 0\\
    \end{bmatrix}$

\vspace{0.7cm}
    \item \textbf{checkerboard(Detecting checkerboard patterns)}

    Checkerboard:
    $\begin{bmatrix}
        1 & -1 & 1 \\
        -1 & 1 & -1 \\
        1 & -1 & 1\\
    \end{bmatrix}$
    \vspace{0.7cm}
    
    \item \textbf{Weights} 

    $Weights={[0.1, 0.1, 0.5, 0.5, 2.0, 0.05]}$
\end{enumerate}

\vspace{1cm}
\noindent\textbf{Note}: We have arrived to the weights given above experimentally.
\subsection{Visualization}
\begin{figure*}[!ht]
\centering
\includegraphics[width=0.98\textwidth]{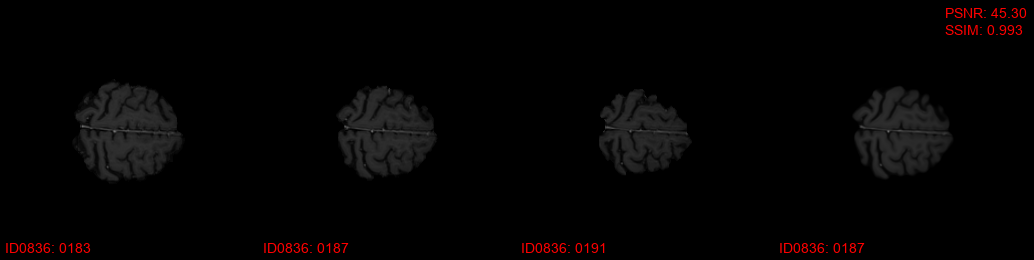}
\caption{ The figure shows the near perfect prediction of sulcal–gyral morphology (folds of the brain surface).
}
\label{fig 5}
\end{figure*}

\begin{figure*}[!ht]
\centering
\includegraphics[width=0.98\textwidth]{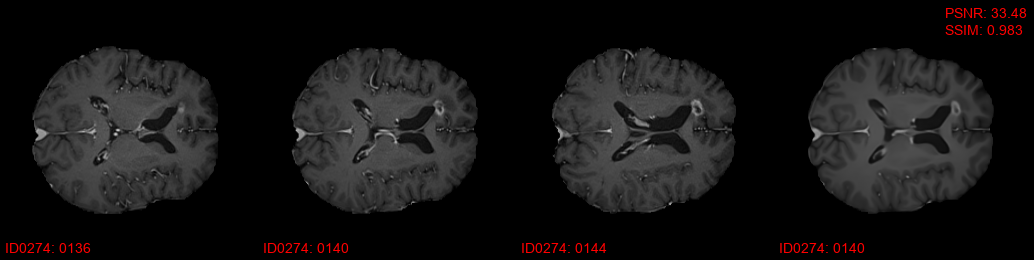}
\caption{ The figure shows the near perfect prediction lesion enhancement (Tumor prediction).
}
\label{fig 6}
\end{figure*}

More detailed qualitative analysis of the 2D slice prediction results is given in this section. The caption for each image explains the image \ref{fig 5},\ref{fig 6}. In particular, Figure \ref{fig 5} shows the effective generation of sulcal-gyral morphology of the brain and Figure \ref{fig 6} shows the near perfect prediction of tumor boundary for a small lesion. Although the PSNR of generated image and the ground truth image is relatively low, the structural consistency is maintained demonstrating the effectiveness of multi-kernel loss and nearly accurate prediction of tumor boundary shows the applicability of the overall model.

\section{3D refinement}
The Kernels and weights for 3D Muti-kernel loss are as follows.

\section*{Kernels and weights for 3D multi-kernel loss}

The following Kernels are \(3\times3\times3\) kernels.  Each Kernels shown as three \(3\times3\) slices (depth slices from front to back).  Let a generic 3D Kernel \(F = [F_{:,:,1},\,F_{:,:,2},\,F_{:,:,3}]\).

\bigskip

\noindent 1. \textbf{Sobel filters (detecting straight edges)}

\begin{equation}
    \resizebox{230 pt}{25 pt}{
    $
    \mathrm{Sobel\_x}=
    \biggl[
\begin{bmatrix}
-1 & 0 & 1\\
-2 & 0 & 2\\
-1 & 0 & 1
\end{bmatrix},\ 
\begin{bmatrix}
-1 & 0 & 1\\
-2 & 0 & 2\\
-1 & 0 & 1
\end{bmatrix},\ 
\begin{bmatrix}
-1 & 0 & 1\\
-2 & 0 & 2\\
-1 & 0 & 1
\end{bmatrix}
\biggr]
$
}
\end{equation}

\begin{equation}
    \resizebox{230 pt}{25 pt}{
    $
    \mathrm{Sobel\_Y}=
\biggl[
\begin{bmatrix}
1 & 2 & 1\\
0 & 0 & 0\\
-1 & -2 & -1
\end{bmatrix},\ 
\begin{bmatrix}
1 & 2 & 1\\
0 & 0 & 0\\
-1 & -2 & -1
\end{bmatrix},\ 
\begin{bmatrix}
1 & 2 & 1\\
0 & 0 & 0\\
-1 & -2 & -1
\end{bmatrix}
\biggr]
$
}
\end{equation}

\begin{equation}
\resizebox{230 pt}{25 pt}{
$
\mathrm{Sobel\_Z}=
\biggl[
\begin{bmatrix}
1 & 1 & 1\\
1 & 1 & 1\\
1 & 1 & 1
\end{bmatrix},\ 
\begin{bmatrix}
0 & 0 & 0\\
0 & 0 & 0\\
0 & 0 & 0
\end{bmatrix},\ 
\begin{bmatrix}
-1 & -1 & -1\\
-1 & -1 & -1\\
-1 & -1 & -1
\end{bmatrix}
\biggr]
$
}
\end{equation}

\bigskip
\noindent 2. \textbf{Laplacian (detecting fine lines and structures)}
\begin{equation}
\resizebox{230 pt}{25 pt}{
$
\mathrm{Laplacian}=
\biggl[
\begin{bmatrix}
0 & 1 & 0\\
1 & -6 & 1\\
0 & 1 & 0
\end{bmatrix},\ 
\begin{bmatrix}
1 & -6 & 1\\
-6 & 24 & -6\\
1 & -6 & 1
\end{bmatrix},\ 
\begin{bmatrix}
0 & 1 & 0\\
1 & -6 & 1\\
0 & 1 & 0
\end{bmatrix}
\biggr]
$
}
\end{equation}

\bigskip
\noindent 3. \textbf{Diagonal filters (detecting diagonal edges)}
\begin{equation}
\resizebox{230 pt}{25 pt}{
$
\mathrm{Diagonal\_1}=
\biggl[
\begin{bmatrix}
0 & 1 & 2\\
-1 & 0 & 1\\
-2 & -1 & 0
\end{bmatrix},\ 
\begin{bmatrix}
0 & 1 & 2\\
-1 & 0 & 1\\
-2 & -1 & 0
\end{bmatrix},\ 
\begin{bmatrix}
0 & 1 & 2\\
-1 & 0 & 1\\
-2 & -1 & 0
\end{bmatrix}
\biggr]
$
}
\end{equation}

\begin{equation}
\resizebox{230 pt}{25 pt}{
$
\mathrm{Diagonal\_2}=
\biggl[
\begin{bmatrix}
2 & 1 & 0\\
1 & 0 & -1\\
0 & -1 & -2
\end{bmatrix},\ 
\begin{bmatrix}
2 & 1 & 0\\
1 & 0 & -1\\
0 & -1 & -2
\end{bmatrix},\ 
\begin{bmatrix}
2 & 1 & 0\\
1 & 0 & -1\\
0 & -1 & -2
\end{bmatrix}
\biggr]
$
}
\end{equation}

\bigskip
\noindent 4. \textbf{Gaussian smoothing (3D separable Gaussian-like stack)}
\begin{equation}
\resizebox{230 pt}{25 pt}{
$
\mathrm{Gaussian}=
\biggl[
\begin{bmatrix}
1 & 2 & 1\\
2 & 4 & 2\\
1 & 2 & 1
\end{bmatrix},\ 
\begin{bmatrix}
2 & 4 & 2\\
4 & 8 & 4\\
2 & 4 & 2
\end{bmatrix},\ 
\begin{bmatrix}
1 & 2 & 1\\
2 & 4 & 2\\
1 & 2 & 1
\end{bmatrix}
\biggr]
$
}
\end{equation}

\bigskip
\noindent 5. \textbf{High-pass edge (sharpen / edge enhancement)}
\begin{equation}
\resizebox{230 pt}{25 pt}{
$
\mathrm{HighPass}=
\biggl[
\begin{bmatrix}
-1 & -1 & -1\\
-1 & 8 & -1\\
-1 & -1 & -1
\end{bmatrix},\ 
\begin{bmatrix}
-1 & -1 & -1\\
-1 & 8 & -1\\
-1 & -1 & -1
\end{bmatrix},\ 
\begin{bmatrix}
-1 & -1 & -1\\
-1 & 8 & -1\\
-1 & -1 & -1
\end{bmatrix}
\biggr]
$
}
\end{equation}

\bigskip
\noindent 6. \textbf{Laplacian of Gaussian (LoG)}
\begin{equation}
\resizebox{230 pt}{25 pt}{
$
\mathrm{LoG}=
\biggl[
\begin{bmatrix}
0 & 0 & -1\\
0 & -1 & -2\\
-1 & -2 & -1
\end{bmatrix},\ 
\begin{bmatrix}
0 & -1 & -2\\
-1 & 16 & -2\\
-2 & -1 & 0
\end{bmatrix},\ 
\begin{bmatrix}
-1 & -2 & -1\\
-2 & -1 & 0\\
0 & 0 & 0
\end{bmatrix}
\biggr]
$
}
\end{equation}

\bigskip
\noindent 7. \textbf{Cross diagonal}
\begin{equation}
\resizebox{230 pt}{25 pt}{
$
\mathrm{CrossDiagonal}=
\biggl[
\begin{bmatrix}
1 & 0 & -1\\
0 & 0 & 0\\
-1 & 0 & 1
\end{bmatrix},\ 
\begin{bmatrix}
1 & 0 & -1\\
0 & 0 & 0\\
-1 & 0 & 1
\end{bmatrix},\ 
\begin{bmatrix}
1 & 0 & -1\\
0 & 0 & 0\\
-1 & 0 & 1
\end{bmatrix}
\biggr]
$
}
\end{equation}

\bigskip
\noindent 8. \textbf{Checkerboard (alternating pattern detector)}
\begin{equation}
\resizebox{230 pt}{25 pt}{
$
\mathrm{Checkerboard}=
\biggl[
\begin{bmatrix}
1 & -1 & 1\\
-1 & 1 & -1\\
1 & -1 & 1
\end{bmatrix},\ 
\begin{bmatrix}
-1 & 1 & -1\\
1 & -1 & 1\\
-1 & 1 & -1
\end{bmatrix},\ 
\begin{bmatrix}
1 & -1 & 1\\
-1 & 1 & -1\\
1 & -1 & 1
\end{bmatrix}
\biggr]
$
}
\end{equation}

\bigskip
\noindent\textbf{Normalization:} each 3D Kernel \(F\) is normalized in the implementation as
\[
\tilde{F}=\frac{F}{\sum_{i,j,k} |F_{i,j,k}| + \epsilon},
\qquad \epsilon=10^{-6}.
\]

\bigskip
\noindent\textbf{Weights:} (assigned importance to each kernel)
\[
\text{weights} = [\,1,\;1,\;1,\;2,\;0.8,\;0.8,\;0.5,\;1.2,\;1.5,\;0.8,\;0.5\,].
\]

\subsection{Radiologist Comments}
The additional comments from the radiologist are as follows:

\noindent Visual inspection of the reconstructed MR images demonstrated that the proposed MK-ResRecon framework substantially enhanced cortical surface continuity and local texture fidelity. On T1-weighted images, mild gyral thickening was observed in the reconstructed volumes relative to the original 2D slices, indicating successful interpolation of inter-slice gaps and preservation of sulcal–gyral morphology. These findings suggest that the multi-kernel residual architecture effectively captures spatial correlations between sampled slices, contributing to smoother and more anatomically consistent cortical transitions.

When evaluated across field strengths, 3 T MRI datasets exhibited superior performance compared with 1.5 T MRI, characterized by sharper cortical boundaries, higher signal-to-noise ratio (SNR), and improved contrast delineation. This suggests that the MK-ResRecon framework benefits from the richer high-frequency information available at higher field strengths, reinforcing its texture-aware volumetric refinement capability.

\end{document}